\documentclass[11pt,a4paper]{article}

\usepackage[hyperref]{acl2018}

\aclfinalcopy 

\usepackage[colorinlistoftodos]{todonotes}

\usepackage{times}
\usepackage{latexsym}

\ifxetex
    \usepackage{fontspec}
\else
    \usepackage[T1]{fontenc}
    \usepackage[utf8]{inputenc}
\fi

\usepackage{url}

\usepackage{graphicx} 
\graphicspath{{images/}}

\usepackage{enumitem}
\usepackage{bbm}
\usepackage{verbatim}
\usepackage{multirow}
\usepackage{tabularx}
\usepackage{booktabs}
\usepackage{pbox}

\usepackage{booktabs}

\usepackage[subtle]{savetrees}

\usepackage{amsfonts}
\usepackage{amssymb}
\usepackage{amsthm}
\usepackage{amsmath}

\DeclareMathOperator{\argmax}{argmax}
\DeclareMathOperator{\softmax}{softmax}
\DeclareMathOperator{\median}{median}

\usepackage{xspace}

\makeatletter
\newcommand*{\etc}{%
	\@ifnextchar{.}%
	{etc}%
	{etc.\@\xspace}%
}
\makeatother

\usepackage{xcolor}

\definecolor{nice-red}{HTML}{E41A1C}
\definecolor{nice-orange}{HTML}{FF7F00}
\definecolor{nice-yellow}{HTML}{FFC020}
\definecolor{nice-green}{HTML}{4DAF4A}
\definecolor{nice-blue}{HTML}{377EB8}
\definecolor{nice-purple}{HTML}{984EA3}

\newcommand{\Real}{\ensuremath{\mathbb{R}}}
\newcommand{\Natural}{\ensuremath{\mathbb{N}}}

\title{Numeracy for Language Models:\\Evaluating and Improving their Ability to Predict Numbers}

\author{Georgios P. Spithourakis \\
  Department of Computer Science  \\
  University College London \\
  {\tt g.spithourakis@cs.ucl.ac.uk} \\
  \And
  Sebastian Riedel \\
  Department of Computer Science  \\
  University College London \\
  {\tt s.riedel@cs.ucl.ac.uk} \\}

\date{}

\begin{document}
\maketitle
\begin{abstract}
Numeracy is the ability to understand and work with numbers.
It is a necessary skill for composing and understanding documents in clinical, scientific, and other technical domains.
In this paper, we explore different strategies for modelling numerals with language models, such as memorisation and digit-by-digit composition,
and propose a novel neural architecture that uses a continuous probability density function to model numerals from an open vocabulary.
Our evaluation on clinical and scientific datasets shows that
using hierarchical models to distinguish numerals from words improves a perplexity metric on the subset of numerals by 2 and 4 orders of magnitude, respectively, over non-hierarchical models.
A combination of strategies can further improve perplexity.
Our continuous probability density function model reduces mean absolute percentage errors by 18\% and 54\% in comparison to the second best strategy for each dataset, respectively.
\end{abstract}

\section{Introduction}

Language models (LMs) are statistical models that assign a probability over sequences of words.
Language models can often help with other tasks, such as
speech recognition~\cite{DBLP:conf/interspeech/MikolovKBCK10,DBLP:conf/interspeech/PrabhavalkarRSL17},
machine translation~\cite{DBLP:conf/conll/LuongKM15,DBLP:journals/csl/GulcehreFXCB17},
text summarisation~\cite{DBLP:conf/emnlp/FilippovaACKV15,DBLP:journals/air/GambhirG17},
question answering~\cite{DBLP:journals/corr/WangYT17},
semantic error detection~\cite{DBLP:conf/bea/ReiY17,DBLP:conf/emnlp/SpithourakisAR16},
and fact checking~\cite{DBLP:conf/emnlp/RashkinCJVC17}.

Numeracy and literacy refer to the ability to comprehend, use, and attach meaning to numbers and words, respectively.
Language models exhibit literacy by being able to assign higher probabilities to sentences that are both grammatical and realistic, as in this example:

\begin{center}
\textit{`I eat an apple'} (grammatical and realistic)

\textit{`An apple eats me'} (unrealistic)

\textit{`I eats an apple'} (ungrammatical)
\end{center}

\begin{figure}[t]
\centering
\includegraphics[width=0.9\columnwidth]{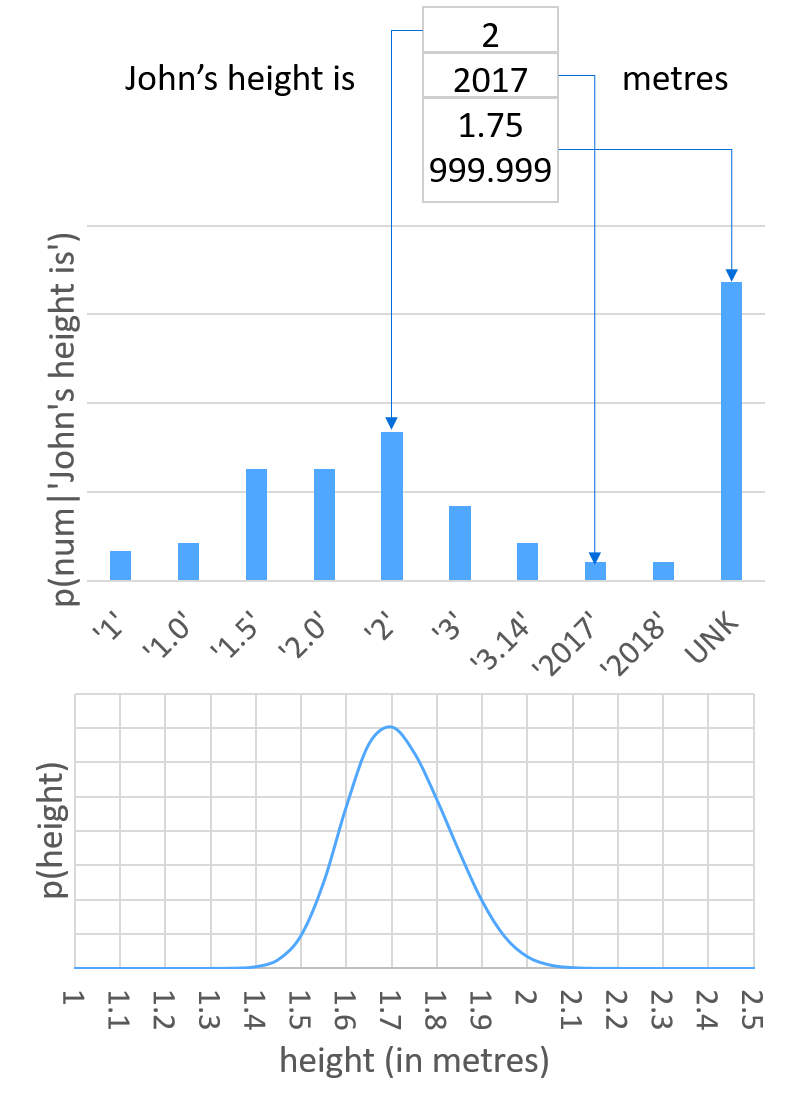}
\caption{Modelling numerals with a categorical distribution over a fixed vocabulary maps all out-of-vocabulary numerals to the same type, e.g. UNK, and does not reflect the smoothness of the underlying continuous distribution of certain attributes.}
\label{fig:motivation}
\end{figure}

Likewise, a numerate language model should be able to rank numerical claims based on plausibility:

\begin{center}
\textit{'John's height is 1.75 metres'} (realistic)
 
\textit{'John's height is 999.999 metres'} (unrealistic)
\end{center}

Existing approaches to language modelling treat numerals similarly to other words, typically using categorical distributions over a fixed vocabulary. However, this maps all unseen numerals to the same unknown type and ignores the smoothness of continuous attributes, as shown in Figure~\ref{fig:motivation}.
In that respect, existing work on language modelling does not explicitly evaluate or optimise for numeracy.
Numerals are often neglected and low-resourced,
e.g. they are often masked~\cite{mitchell2009language},
and there are only $15,164$ $(3.79\%)$ numerals among GloVe's $400,000$ embeddings pretrained on 6 billion tokens~\cite{pennington2014glove}.
Yet, numbers appear ubiquitously, from children's magazines \cite{joram1995numeracy} to clinical reports~\cite{bigeard2015automatic}, and grant objectivity to sciences~\cite{porter1996trust}.


Previous work finds that numerals have higher out-of-vocabulary rates than other words and proposes solutions for representing unseen numerals as inputs to language models, e.g. using numerical magnitudes as features~\cite{spithourakis2016clinical,DBLP:conf/emnlp/SpithourakisAR16}. Such work identifies that the perplexity of language models on the subset of numerals can be very high, but does not directly address the issue.
This paper focuses on evaluating and improving the ability of language models to predict numerals. The main contributions of this paper are as follows:
\begin{enumerate}
   \item We explore different strategies for modelling numerals, such as memorisation and digit-by-digit composition, and propose a novel neural architecture based on continuous probability density functions.
   \item We propose the use of evaluations that adjust for the high out-of-vocabulary rate of numerals and account for their numerical value (magnitude).
   \item We evaluate on a clinical and a scientific corpus and provide a qualitative analysis of learnt representations and model predictions. We find that modelling numerals separately from other words can drastically improve the perplexity of LMs, that different strategies for modelling numerals are suitable for different textual contexts, and that continuous probability density functions can improve the LM's prediction accuracy for numbers.
\end{enumerate}

\section{Language Models}

Let $s_1, s_2, ..., s_L$ denote a document, where $s_t$ is the token at position $t$.
A language model estimates the probability of the next token given previous tokens,
i.e. $p(s_t|s_1, ..., s_{t-1})$.
Neural LMs estimate this probability by feeding embeddings, i.e. vectors that represent each token, into a Recurrent Neural Network (RNN)~\cite{DBLP:conf/interspeech/MikolovKBCK10}.

\paragraph{Token Embeddings}
Tokens are most commonly represented by a $D$-dimensional dense vector
that is unique for each word from a vocabulary $\mathcal{V}$ of known words.
This vocabulary includes special symbols (e.g. `UNK') to handle out-of-vocabulary tokens, such as unseen words or numerals.
Let $w_{s}$ be the one-hot representation of token $s$,
i.e. a sparse binary vector with a single element set to $1$ for that token's index in the vocabulary,
and $E \in \Real^{D \times |\mathcal{V}|}$ be the token embeddings matrix.
The token embedding for $s$ is the vector $e^{\text{token}}_s=Ew_{s}$.

\paragraph{Character-Based Embeddings}
A representation for a token can be build from its constituent characters~\cite{luong2016achieving,santos2014learning}.
Such a representation takes into account the internal structure of tokens.
Let $d_1,d_2,...,d_N$ be the characters of token $s$.
A character-based embedding for $s$ is the final hidden state of a $D$-dimensional character-level RNN: $e^{\text{chars}}_s=\text{RNN}\left( d_0, d_1, ... d_L \right)$.


\paragraph{Recurrent and Output Layer}
The computation of the conditional probability of the next token involves recursively feeding
the embedding of the current token $e_{s_t}$
and the previous hidden state $h_{t-1}$
into a $D$-dimensional token-level RNN
to obtain the current hidden state $h_{t}$.
The output probability is estimated using the softmax function,
i.e.
\begin{equation}
\begin{array}{c}
p \left(s_t \right | h_t)
 = \softmax \left(\psi (s_t) \right)
= \frac{1}{Z} e^{ \psi \left( s_t \right) } \\
Z  = \sum\limits_{s' \in \mathcal{V}}  e^{ \psi (s') },
\end{array}
\label{eq:softmax}
\end{equation}
where $\psi \left(. \right) $ is a score function.

\paragraph{Training and Evaluation}
Neural LMs are typically trained to minimise the cross entropy on the training corpus:
\begin{equation}
\mathcal{H}_{train} =
- \frac{1}{N} \sum_{s_t \in train} \log p \left(s_t \right | s_{<t}) 
\end{equation}

A common performance metric for LMs is per token perplexity (Eq.~\ref{eq:pp}), evaluated on a test corpus. It can also be interpreted as the branching factor: the size of an equally weighted distribution with equivalent uncertainty, i.e. how many sides you need on a fair die to get the same uncertainty as the model distribution.
\begin{equation}
PP_{test} = \exp (\mathcal{H}_{test})
\label{eq:pp}
\end{equation}


\section{Strategies for Modelling Numerals}


In this section we describe models with different strategies for generating numerals and propose the use of number-specific evaluation metrics that adjust for the high out-of-vocabulary rate of numerals and account for numerical values.
We draw inspiration from theories of numerical cognition.
The triple code theory~\cite{dehaene2003three} postulates that humans process quantities through two exact systems (verbal and visual) and one approximate number system that semantically represents a number on a mental number line.
\newcite{tzelgov2015primitives} identify two classes of numbers: i) primitives, which are holistically retrieved from long-term memory; and ii) non-primitives, which are generated online.
An in-depth review of numerical and mathematical cognition can be found in \newcite{kadosh2015oxford} and \newcite{campbell2005handbook}.


\subsection{Softmax Model and Variants}

This class of models assumes that numerals come from a finite vocabulary that can be memorised and retrieved later.
The \emph{softmax} model treats all tokens (words and numerals) alike and directly uses Equation~\ref{eq:softmax} with score function:
\begin{equation}
\psi(s_t)
= h^T_t e^{\text{token}}_{s_t}
= h^T_tE_{\text{out}}w_{s_t},
\end{equation}
where
$E_{out} \in \Real^{D \times |\mathcal{V}|}$
is an output embeddings matrix.
The summation in Equation~\ref{eq:softmax} is over the complete target vocabulary,
which requires mapping any out-of-vocabulary tokens to special symbols, e.g. `$\text{UNK}_\text{word}$' and `$\text{UNK}_\text{numeral}$'.


\paragraph{Softmax with Digit-Based Embeddings}
The \emph{softmax+rnn} variant considers the internal syntax of a numeral's digits
by adjusting the score function:
\begin{equation}
\begin{split}
\psi (s_t) & = h^T_t e^{\text{token}}_{s_t} + h^T_t e^{\text{chars}}_{s_t} \\
& = h^T_tE_{\text{out}}w_{s_t} + h^T_tE^{\text{RNN}}_{\text{out}}w_{s_t},    
\end{split}
\end{equation}
where the columns of $E^{\text{RNN}}_{\text{out}}$ are composed of character-based embeddings for in-vocabulary numerals and token embeddings for the remaining vocabulary.
The character set comprises digits (0-9), the decimal point, and an end-of-sequence character.
The model still requires normalisation over the whole vocabulary,
and the special unknown tokens are still needed.

\paragraph{Hierarchical Softmax}
A hierarchical softmax~\cite{DBLP:conf/aistats/MorinB05} can help us decouple the modelling of numerals from that of words.
The probability of the next token $s_t$ is decomposed to that of its class $c_t$
and the probability of the exact token from within the class:
\begin{equation}
\begin{array}{c}
p(s_t|h_t) = \sum\limits_{c_t \in C} p(c_t|h_t)p(s_t|c_t,h_t)  \\
p(c_t|h_t) = \sigma \left( h^T_t b\right)\\
\end{array}
\end{equation}
where
the valid token classes are $C=\{ \text{word, numeral} \}$,
$\sigma$ is the sigmoid function and $b$ is a $D$-dimensional vector.
Each of the two branches of $p(s_t|c_t,h_t)$ can now be modelled by independently normalised distributions.
The hierarchical variants (\emph{h-softmax} and \emph{h-softmax+rnn}) use two independent softmax distributions for words and numerals. The two branches share no parameters, and thus words and numerals will be embedded into separate spaces.

The hierarchical approach allows us to use any well normalised distribution to model each of its branches.
In the next subsections, we examine different strategies for modelling the branch of numerals,
i.e. $p(s_t|c_t=\text{numeral},h_t)$. For simplicity, we will abbreviate this to $p(s)$.

\subsection{Digit-RNN Model}

Let $d_1,d_2...d_N$ be the digits of numeral $s$.
A digit-by-digit composition strategy estimates the probability of the numeral from the probabilities of its digits:
\begin{equation}
p(s) = p(d_1)p(d_2|d_1)...p(d_N|d_{<N})
\end{equation}
The \emph{d-RNN} model feeds the hidden state $h_t$ of the token-level RNN into a character-level RNN~\cite{graves2013generating,sutskever2011generating} to estimate this probability.
This strategy can accommodate an open vocabulary, i.e. it eliminates the need for an $\text{UNK}_\text{numeral}$ symbol,
as the probability is normalised one digit at a time over the much smaller vocabulary of digits (digits 0-9, decimal separator, and end-of-sequence).

\subsection{Mixture of Gaussians Model}

Inspired by the approximate number system and the mental number line~\cite{dehaene2003three},
our proposed \emph{MoG} model computes the probability of numerals from a probability density function (pdf) over real numbers,
using a mixture of Gaussians for the underlying pdf:
\begin{equation}
\begin{split}
q(v) &= \sum_{k=1}^K  \pi_k \mathcal{N}_k(v; \mu_k,\sigma_k^2) \\
\pi_k &= \softmax \left( B^T h_t \right),
\end{split}
\end{equation}
where $K$ is the number of components,
$\pi_k$ are mixture weights that depend on hidden state $h_t$ of the token-level RNN,
$\mathcal{N}_k$ is the pdf of the normal distribution with mean $\mu_k \in \Real$ and variance $\sigma_k^2 \in \Real $,
and $B\in \Real^{D \times K}$ is a matrix.

The difficulty with this approach is that for any continuous random variable, the probability that it equals a specific value is always zero.
To resolve this, we consider a probability mass function (pmf) that discretely approximates the pdf:
\begin{equation}
\widetilde{Q}(v|r) = \int\displaylimits_{v - \epsilon_r}^{v + \epsilon_r} q(u) du
 = 
 F \left( v + \epsilon_r \right)
 - F \left( v -  \epsilon_r \right) ,
\end{equation}
where $F(.)$ is the cumulative density function of $q(.)$,
and $\epsilon_r = 0.5 \times 10^{-r}$ is the number's precision.
The level of discretisation $r$,
i.e. how many decimal digits to keep,
is a random variable in $\Natural$ with distribution $p(r)$. The mixed joint density is:
\begin{equation}
p(s)=p(v,r)=p(r)\widetilde{Q}(v|r)
\end{equation}

\begin{figure}[t]
\centering
\includegraphics[width=0.90\columnwidth]{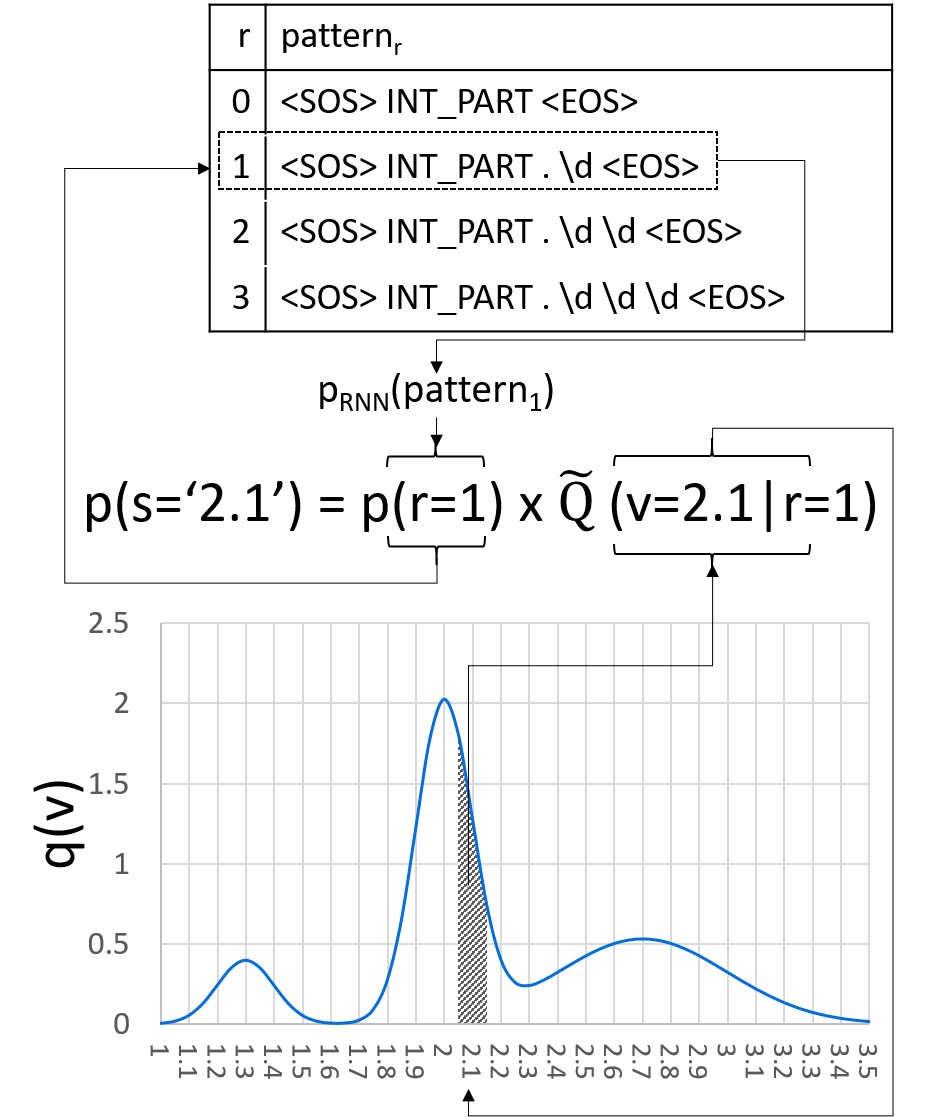}
\caption{Mixture of Gaussians model.
The probability of a numeral is decomposed into the probability of its decimal precision and the probability that an underlying number will produce the numeral when rounded at the given precision.}
\label{fig:mode_continuous}
\end{figure}

Figure~\ref{fig:mode_continuous} summarises this strategy,
where we model the level of discretisation by converting the numeral into a pattern and use a RNN to estimate the probability of that pattern sequence:
\begin{equation}
p(r)=p(\text{SOS INT\_PART . }\overbrace{\setminus\text{d ... }\setminus\text{d}}^{r \text{ decimal digits}}\text{ EOS})
\end{equation}



\subsection{Combination of Strategies}

Different mechanisms might be better for predicting numerals in different contexts.
We propose a \emph{combination} model that can select among different strategies for modelling numerals:
\begin{equation}
\begin{split}
p(s) & = \sum_{\forall m \in M} \alpha_m p(s|m) \\
\alpha_m & = \softmax \left( A^Th_t \right),
\end{split}
\end{equation}
where M=\{h-softmax, d-RNN, MoG\},
and $A \in \Real^{D \times |M|}$.
Since both d-RNN and MoG are open-vocabulary models, the unknown numeral token can now be removed from the vocabulary of h-softmax.


\subsection{Evaluating the Numeracy of LMs}


Numeracy skills are centred around the understanding of numbers and numerals.
A number is a mathematical object with a specific magnitude,
whereas a numeral is its symbolic representation,
usually in the positional decimal Hindu–Arabic numeral system~\cite{mccloskey1995representing}.
In humans, the link between numerals and their numerical values boosts numerical skills~\cite{griffin1995teaching}.

\paragraph{Perplexity Evaluation}
Test perplexity evaluated only on numerals will be informative of the symbolic component of numeracy.
However, model comparisons based on naive evaluation using Equation~\ref{eq:pp} might be problematic:
perplexity is sensitive to out-of-vocabulary (OOV) rate, which might differ among models, e.g. it is zero for open-vocabulary models.
As an extreme example, in a document where all words are out of vocabulary, the best perplexity is achieved by a trivial model that predicts everything as unknown.

~\newcite{ueberla1994analysing} proposed Adjusted Perplexity (APP; Eq.~\ref{eq:app}),
also known as unknown-penalised perplexity~\cite{ahn2016neural},
to cancel the effect of the out-of-vocabulary rate on perplexity.
The APP is the perplexity of an adjusted model
that uniformly redistributes the probability of each out-of-vocabulary class over all different types in that class:
\begin{equation}
p'(s) = 
\begin{cases}
      p(s)  \frac{1}{|OOV_c|} & \text{if } s \in OOV_c \\
      p(s)   & \text{otherwise} \\
\end{cases}
\end{equation}
where $OOV_{c}$ is an out-of-vocabulary class (e.g. words and numerals),
and $|OOV_c|$ is the cardinality of each OOV set.
Equivalently, adjusted perplexity can be calculated as:
\begin{equation}
\begin{split}
APP_{test} = \exp \left( \mathcal{H}_{test} +  \sum_{c}\mathcal{H}^c_{\text{adjust}} \right) \\
\mathcal{H}_{\text{adjust}}^c = - \sum_{t} \frac{|s_t \in OOV_c|}{N} \log{\frac{1}{|OOV_c|}}
\end{split}
\label{eq:app}
\end{equation}
where $N$ is the total number of tokens in the test set
and
$|s \in OOV_c|$ is the count of tokens from the test set belonging in each OOV set.

\paragraph{Evaluation on the Number Line}
While perplexity looks at symbolic performance on numerals, this evaluation focuses on numbers and particularly on their numerical value, which is their most prominent semantic content~\cite{dehaene2003three, dehaene1995towards}.

Let $v_t$ be the numerical value of token $s_t$ from the test corpus.
Also, let $\hat{v}_t$ be the value of the most probable numeral under the model $s_t=\argmax \left( p(s_t|h_t, c_t=\text{num}) \right)$.
Any evaluation metric from the regression literature can be used to measure the models performance.
To evaluate on the number line, we can use any evaluation metric from the regression literature.
In reverse order of tolerance to extreme errors, some of the most popular are
Root Mean Squared Error (RMSE),
Mean Absolute Error (MAE),
and Median Absolute Error (MdAE):

\begin{equation}
\begin{array}{rcl}
e_{i} &=& v_i-\hat{v_i} \\
RMSE &=& \sqrt{\frac{1}{N}\sum\limits_{i=1}^{N}e_i^2} \\
MAE  &=& \frac{1}{N}\sum\limits_{i=1}^{N}|e_i| \\
MdAE &=& \median \{ |e_i| \}  
\end{array}
\label{eq:rmse}
\end{equation}

The above are sensitive to the scale of the data.
If the data contains values from different scales,
percentage metrics are often preferred,
such as the
Mean/Median Absolute Percentage Error (MAPE/MdAPE):

\begin{equation}
\begin{array}{rcl}
pe_{i} & = & \frac{v_i-\hat{v_i}}{v_i} \\
MAPE &=& \frac{1}{N}\sum\limits_{i=1}^{N}{\left| pe_i \right|} \\
MdAPE &=& \median \{ |pe_i| \}
\end{array}
\label{eq:mape}
\end{equation}


\section{Data}

To evaluate our models, we created two datasets with documents from the clinical and scientific domains, where numbers abound~\cite{bigeard2015automatic,porter1996trust}.
Furthermore, to ensure that the numbers will be informative of some attribute,
we only selected texts that reference tables.

\paragraph{Clinical Data}
Our \emph{clinical} dataset comprises clinical records from the London Chest Hospital.
The records where accompanied by tables with 20 numeric attributes (age, heart volumes, etc.) that they partially describe, as well as include numbers not found in the tables.
Numeric tokens constitute only a small proportion of each sentence (4.3\%), but account for a large part of the unique tokens vocabulary (>40\%) and suffer high OOV rates.

\paragraph{Scientific Data}
Our \emph{scientific} dataset comprises paragraphs from Cornell’s ARXIV~\footnote{ARXIV.ORG. Cornell University Library at http://arxiv.org/, visited December 2016} repository of scientific articles, with more than half a million converted papers in 37 scientific sub-fields.
We used the preprocessed ARXMLIV~\cite{stamerjohanns2010transforming,stamerjohanns2008transforming}~\footnote{ARXMLIV. Project home page at http://arxmliv.kwarc.info/, visited December 2016} version, where papers have been converted from LATEX into a custom XML format using the LATEXML~\footnote{LATEXML. http://dlmf.nist.gov, visited December 2016} tool. We then kept all paragraphs with at least one reference to a table and a number.

\begin{table}[h]
\begin{center}
\resizebox{\columnwidth}{!}{%
\begin{tabular}{rrrrrrr}
\toprule    
 &              & \textbf{Clinical} &             &           & \textbf{Scientific} & \\
 & \textbf{Train} & \textbf{Dev} & \textbf{Test}  & \textbf{Train} & \textbf{Dev} & \textbf{Test} \\

\midrule

\#inst   &  11170 &    1625 &   3220  &  14694 &    2037 &   4231  \\
maxLen  &    667 &    594 &     666   &   2419 &    1925 &   1782  \\
avgLen & 210.1 &  209.1 & 206.9   & 210.1 &  215.9 & 212.1  \\
\%word     &  95.7 &   95.7 &  95.7   &  96.1 &   96.1 &  96.0  \\
\%nums  &   4.3 &    4.3 &   4.3    &   3.9 &    3.9 &   4.0  \\

\cmidrule(lr){1-1}
\cmidrule(lr){2-4}
\cmidrule(lr){5-7}

min     &       0.0 &      0.0 &       0.0    &  0.0 & 0.0 & 0.0 \\
median  &      59.5 &     59.0 &      60.0  &  5.0 & 4.0 & 4.5 \\
mean    &     300.6 &    147.7 &     464.8  &  $\sim10^{21}$ & $\sim10^{7}$ & $\sim10^{7}$ \\
max     &  $\sim10^{7}$ & $\sim10^{5}$ & $\sim10^{7}$  &  $\sim10^{26}$ & $\sim10^{11}$ & $\sim10^{11}$ \\
\bottomrule
\end{tabular}
}
\end{center}
\caption{Statistical description of the clinical and scientific datasets: Number of instances (i.e. paragraphs), maximum and average lengths, proportions of words and numerals, descriptive statistics of numbers.}
\label{tab:describe_data}
\end{table}

For both datasets, we lowercase tokens and normalise numerals by omitting the thousands separator ("2,000" becomes "2000") and leading zeros ("007" becomes "7").
Special mathematical symbols are tokenised separately, e.g. negation (``-1'' as ``-'', ``1''), fractions (``3/4'' as ``3'', ``/'', ``4''), etc. For this reason, all numbers were non-negative. Table~\ref{tab:describe_data} shows descriptive statistics for both datasets.

\section{Experimental Results and Discussion}

\begin{table*}[t]
\begin{center}
\resizebox{\linewidth}{!}{%
\begin{tabular}{rrrrrrrrrrrrr}
\toprule 
&\multicolumn{6}{c}{\textbf{Clinical}} & \multicolumn{6}{c}{\textbf{Scientific}} \\

\cmidrule(lr){2-7}
\cmidrule(lr){8-13}

 & \multicolumn{2}{c}{\textbf{words}} &\multicolumn{2}{c}{\textbf{numerals}} &\multicolumn{2}{c}{\textbf{total}}
 & \multicolumn{2}{c}{\textbf{words}} &\multicolumn{2}{c}{\textbf{numerals}} & \multicolumn{2}{c}{\textbf{total}}
\\
\textbf{Model} & \textbf{PP} & \textbf{APP} & \textbf{PP} & \textbf{APP} & \textbf{PP} & \textbf{APP} & \textbf{PP} & \textbf{APP} & \textbf{PP} & \textbf{APP} & \textbf{PP} & \textbf{APP}\\

\cmidrule(lr){1-1}
\cmidrule(lr){2-3}
\cmidrule(lr){4-5}
\cmidrule(lr){6-7}
\cmidrule(lr){8-9}
\cmidrule(lr){10-11}
\cmidrule(lr){12-13}

softmax  &  4.08 &     5.99   & 12.04  & 58443.72 &  4.28 &     8.91 
        &  33.96 &      51.83 & 127.12 & 3505856.25  & 35.79 &      80.62
           \\

softmax+rnn     &  4.03 &     5.91  & \textbf{11.57} & 56164.81 & 4.21 &     8.77 
                &  \textbf{33.54} &      51.20 & \textbf{119.68} & 3300688.50  & \textbf{35.28} &      79.47 
            \\
            
h-softmax  &  \textbf{4.00} &   4.96  & 11.78 & 495.95  &  \textbf{4.19}  &   6.05   
             &  34.73 &  49.81 & 122.67 & 550.98 &  36.51 &  54.80 
            \\
            
h-softmax+rnn     &  4.03 &   4.99  & 11.65 & 490.14 &  4.22 &   6.09
                  &  34.04 &   48.83 & 120.83 &  542.70  &  35.80 &   53.73
                \\
                
d-RNN           &   3.99 &   \textbf{4.95}   & 263.22 & 263.22  &   4.79 &   5.88
           &  34.08 &  48.89  & 519.80 & \textbf{519.80}  &  37.98 &  53.70 
            \\

MoG      &   4.03 &   4.99 & 226.46 & 226.46    &   4.79 &   5.88  
           &  34.14 &  48.97  & 683.16 & 683.16 &  38.45 &  54.37
        \\
        
combination      &   4.01  &   4.96  & 197.59  & \textbf{197.59}   &   4.74  &   \textbf{5.82}
               &  33.64 &  \textbf{48.25}  & 520.95 & 520.95 & 37.50 &  \textbf{53.03}
              \\
\bottomrule
\end{tabular}
}
\end{center}
\caption{Test set perplexities for the clinical and scientific data.
Adjusted perplexities (APP) are directly comparable across all data and models,
but perplexities (PP) are sensitive to the varying out-of-vocabulary rates.}
\label{tab:results_pp}
\end{table*}

\begin{table*}[t]
\begin{center}
\resizebox{\linewidth}{!}{%
\begin{tabular}{rrrrrrrrr}
\toprule
&\multicolumn{5}{c}{\textbf{Clinical}} & \multicolumn{3}{c}{\textbf{Scientific}} \\
\textbf{Model} & \textbf{RMSE} & \textbf{MAE} & \textbf{MdAE} & \textbf{MAPE\%} & \textbf{MdAPE\%} & \textbf{MdAE} & \textbf{MAPE\%} & \textbf{MdAPE\%} \\

\cmidrule(lr){1-1}
\cmidrule(lr){2-6}
\cmidrule(lr){7-9}

mean
& 1043.68 & 294.95 & 245.59 & 2353.11 & 409.47
& $\sim 10^{20}$ & $\sim 10^{23}$ & $\sim 10^{22}$
\\

median
& 1036.18 & 120.24 & 34.52 & 425.81 & 52.05
& 4.20 & 8039.15 & 98.65
\\

\cmidrule(lr){1-1}
\cmidrule(lr){2-6}
\cmidrule(lr){7-9}

softmax
& 997.84 & 80.29 & 12.70 & 621.78 & 22.41
& 3.00 & 1947.44 & 80.62
\\

softmax+rnn
& 991.38 & 74.44 & 13.00 & 503.57 & 23.91
& 3.50 & 15208.37 & 80.00
\\

h-softmax
& 1095.01 & 167.19 & 14.00 & 746.50 & 25.00
& 3.00 & 1652.21 & 80.00
\\

h-softmax+rnn
& 1001.04 & 83.19 & 12.30 & 491.85 & 23.44
& 3.00 & 2703.49 & 80.00
\\

d-RNN
& 1009.34 & 70.21 & 9.00 &  513.81 &  17.90
& 3.00 & 1287.27 & \textbf{52.45}  
\\

MoG
& 998.78 & \textbf{57.11} & \textbf{6.92} &  \textbf{348.10} & \textbf{13.64}
& \textbf{2.10} & \textbf{590.42} & 90.00
\\

combination
& \textbf{989.84} & 69.47 & 9.00  & 552.06 & 17.86
& 3.00 & 2332.50 & 88.89
\\

\bottomrule
\end{tabular}
}%
\end{center}
\caption{Test set regression evaluation for the clinical and scientific data. Mean absolute percentage error (MAPE) is scale independent and allows for comparison across data, whereas root mean square and mean absolute errors (RMSE, MAE) are scale dependent. Medians (MdAE, MdAPE) are informative of the distribution of errors.}
\label{tab:results_reg}
\end{table*}

We set the vocabularies to the 1,000 and 5,000 most frequent token types for the clinical and scientific datasets, respectively. We use gated token-character embeddings~\cite{miyamoto2016gated} for the input of numerals and token embeddings for the input and output of words, since the scope of our paper is numeracy.
We set the models' hidden dimensions to $D=50$ and initialise all token embeddings to pretrained GloVe~\cite{pennington2014glove}.
All our RNNs are LSTMs~\cite{hochreiter1997long} with the biases of LSTM forget gate were initialised to 1.0~\cite{DBLP:conf/icml/JozefowiczZS15}.
We train using mini-batch gradient decent with the Adam optimiser~\cite{kingma2014adam} and regularise with early stopping and $0.1$ dropout rate~\cite{srivastava2013improving} in the input and output of the token-based RNN.

For the mixture of Gaussians, we select the mean and variances to summarise the data at different granularities by fitting 7 separate mixture of Gaussian models on all numbers, each with twice as many components as the previous, for a total of $2^{7+1}-1=256$ components. These models are initialised at percentile points from the data and trained with the expectation-minimisation algorithm. The means and variances are then fixed and not updated when we train the language model.


\subsection{Quantitative Results}

\paragraph{Perplexities}
Table~\ref{tab:results_pp} shows perplexities evaluated on the subsets of words, numerals and all tokens of the test data.
Overall, all models performed better on the clinical than on the scientific data.
On words, all models achieve similar perplexities in each dataset.

On numerals, softmax variants perform much better than other models in PP, which is an artefact of the high OOV-rate of numerals. APP is significantly worse, especially for non-hierarchical variants,
which perform about 2 and 4 orders of magnitude worse than hierarchical ones.

For open-vocabulary models, i.e. d-RNN, MoG, and combination, PP is equivalent to APP.
On numerals, d-RNN performed better than softmax variants in both datasets.
The MoG model performed twice as well as softmax variants on the clinical dataset, but had the third worse performance in the scientific dataset.
The combination model had the best overall APP results for both datasets.

\paragraph{Evaluations on the Number Line}

To factor out model specific decoding processes for finding the best next numeral, we use our models to rank a set of candidate numerals: we compose the union of in-vocabulary numbers and 100 percentile points from the training set, and we convert numbers into numerals by considering all formats up to $n$ decimal points. We select $n$ to represent 90\% of numerals seen at training, which yields $n=3$ and $n=4$ for the clinical and scientific data, respectively.

Table~\ref{tab:results_reg} shows evaluation results, where we also include two naive baselines of constant predictions: with the mean and median of the training data.
For both datasets, RMSE and MAE were too sensitive to extreme errors to allow drawing safe conclusions, particularly for the scientific dataset,
where both metrics were in the order of $10^9$.
MdAE can be of some use, as 50\% of the errors are absolutely smaller than that.

Along percentage metrics, MoG achieved the best MAPE in both datasets (18\% and 54\% better that the second best) and was the only model to perform better than the median baseline for the clinical data.
However, it had the worst MdAPE, which means that MoG mainly reduced larger percentage errors.
The d-RNN model came third and second in the clinical and scientific datasets, respectively.
In the latter it achieved the best MdAPE, i.e. it was effective at reducing errors for 50\% of the numbers.
The combination model did not perform better than its constituents.
This is possibly because MoG is the only strategy that takes into account the numerical magnitudes of the numerals.


\subsection{Learnt Representations}

\begin{figure}[ht]
\centering
\includegraphics[width=0.7\columnwidth]{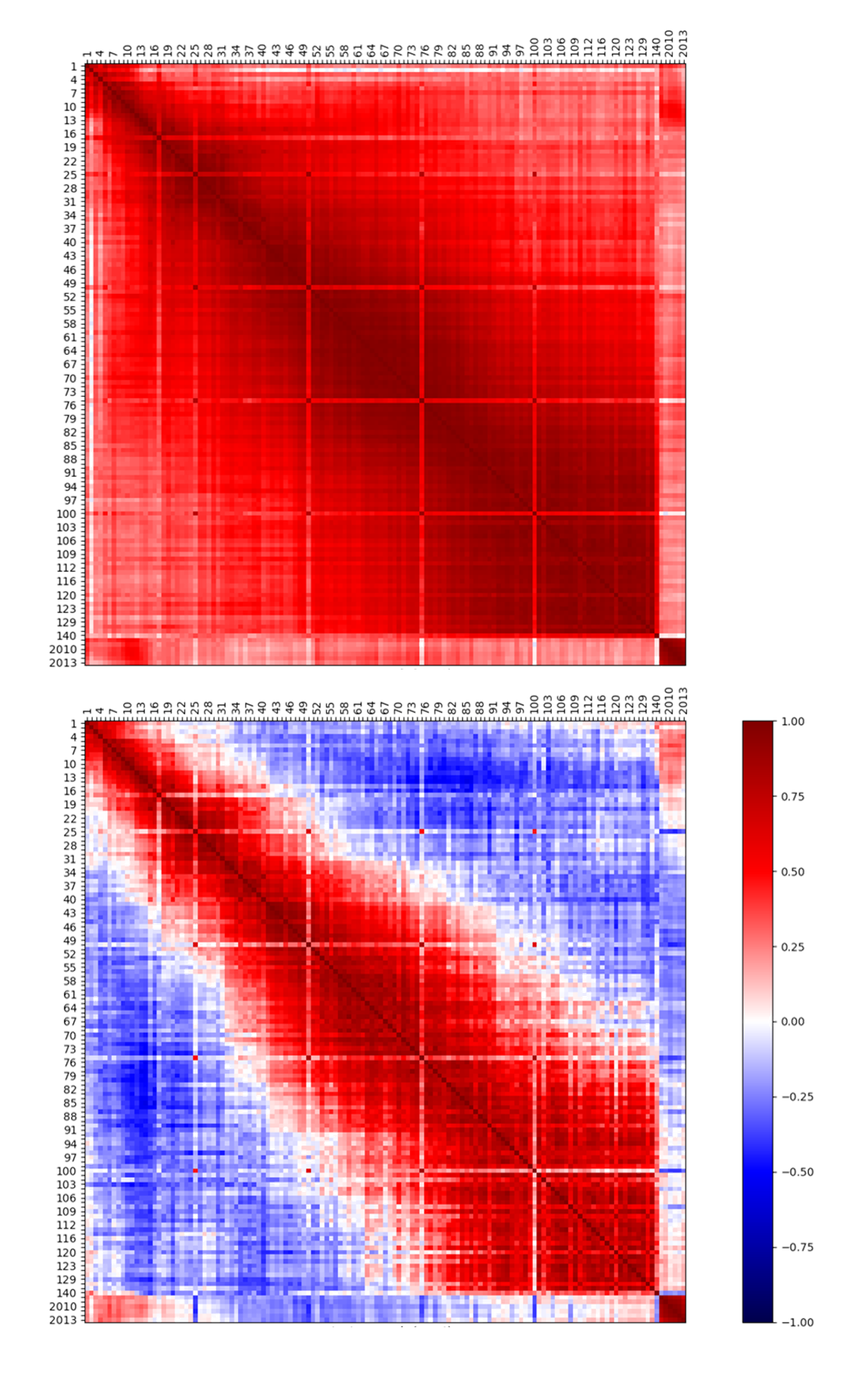}
\caption{Numeral embeddings for the softmax (top) and h-softmax (bottom) models on the clinical data. Numerals are sorted by value.}
\label{fig:softmax_problem}
\end{figure}
\paragraph{Softmax versus Hierarchical Softmax}
Figure~\ref{fig:softmax_problem} visualises the cosine similarities of the output token embeddings of numerals for the softmax and h-softmax models.
Simple softmax enforced high similarities among all numerals and the unknown numeral token,
so as to make them more dissimilar to words,
since the model embeds both in the same space.
This is not the case for h-softmax that uses two different spaces:
similarities are concentrated along the diagonal and fan out as the magnitude grows,
with the exception of numbers with special meaning,
e.g. years and percentile points.

\begin{figure}[ht]
\centering
\includegraphics[width=0.62\columnwidth]{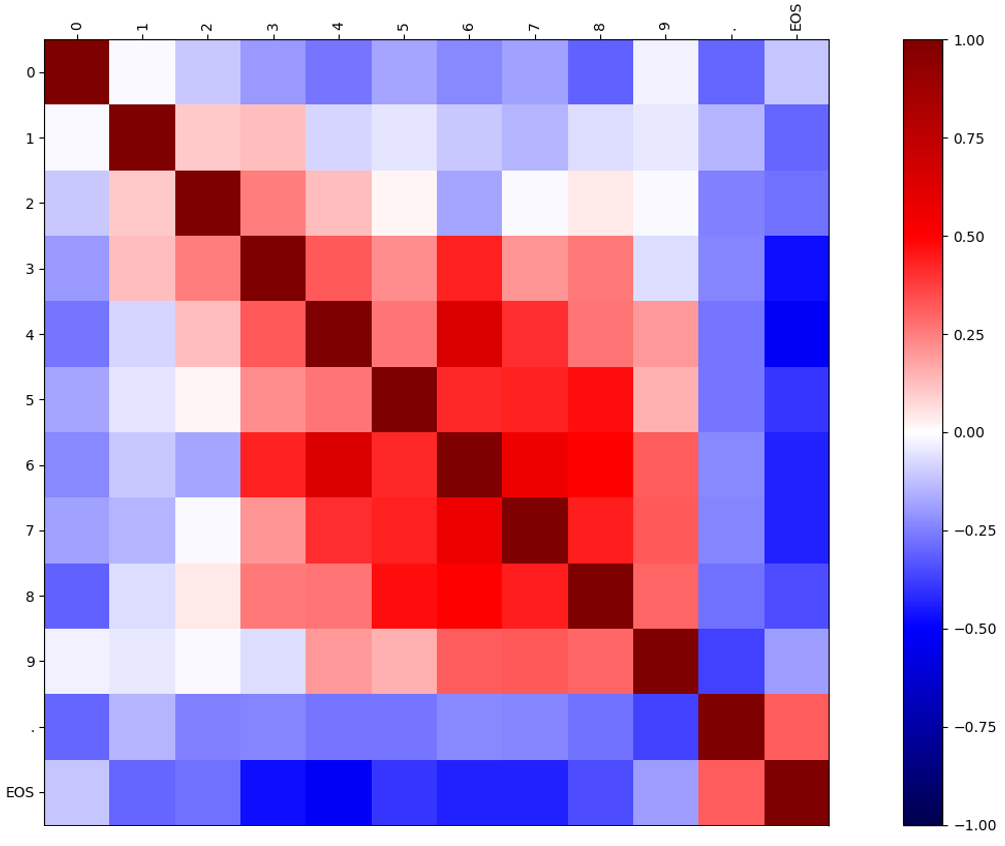}
\caption{Cosine similarities for d-RNN's output digit embeddings trained on the scientific data.}
\label{fig:digits}
\end{figure}

\paragraph{Digit embeddings}
Figure~\ref{fig:digits} shows the cosine similarities between the digits of the d-RNN output mode.
We observe that each primitive digit is mostly similar to its previous and next digit. Similar behaviour was found for all digit embeddings of all models.


\subsection{Predictions from the Models}

\begin{figure}[ht]
\centering
\includegraphics[width=0.98\columnwidth]{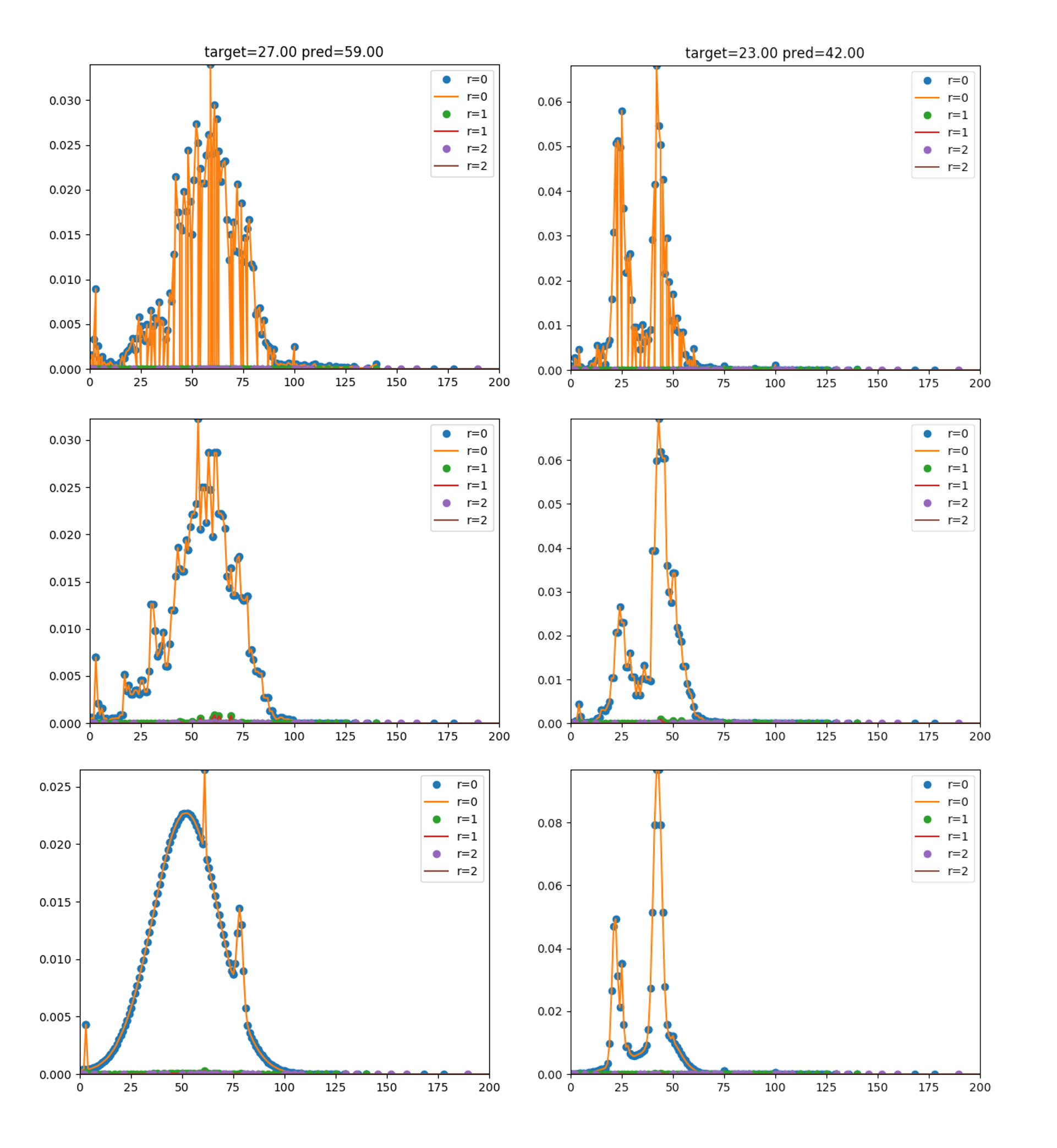}
\caption{Example model predictions for the h-softmax (top), d-RNN (middle) and MoG (bottom) models. Examples from the clinical development set.}
\label{fig:pdf}
\end{figure}

\paragraph{Next Numeral}
Figure~\ref{fig:pdf} shows the probabilities of different numerals under each model for two examples from the clinical development set. Numerals are grouped by number of decimal points. The h-softmax model's probabilities are spiked, d-RNNs are saw-tooth like and MoG's are smooth, with the occasional spike, whenever a narrow component allows for it. Probabilities rapidly decrease for more decimal digits, which is reminiscent of the theoretical expectation that the probability of en exact value for a continuous variable is zero.

\begin{table*}[t]
\begin{center}
\begin{tabular}{cll}
\toprule
 & \multicolumn{1}{c}{\textbf{Clinical}} & \multicolumn{1}{c}{\textbf{Scientific}} \\
\midrule
\parbox[t]{2mm}{\rotatebox[origin=c]{90}{\textbf{h-softmax}}}
&
\pbox{7.7cm}{
\textbf{Examples}:
\textit{
``late enhancement ( $>$ \textbf{75} \%)'',
``late gadolinium enhancement ( $<$ \textbf{25} \%)'',
``infarction ( \textbf{2} out of \textbf{17} segments )'',
``infarct with \textbf{4} out of \textbf{17} segments nonviable'',
``adenosine stress perfusion @ \textbf{140} mcg'',
``stress perfusion ( adenosine \textbf{140} mcg''
}\\
\textbf{Numerals}: 50, 17, 100, 75, 25, 1, 140, 2012, 2010, 2011, 8, 5, 2009, 2013, 7, 6, 2, 3, 2008, 4...
}
&
\pbox{6cm}{
\textbf{Examples}:
\textit{
``sharp et al . \textbf{2004}'',
``li et al . \textbf{2003}'',
``3.5 $\times$ \textbf{10} $\hat{}$ 4'',
``0.3 $\times$ \textbf{10} $\hat{}$ 16''
} \\
\textbf{Numerals}:
1992, 2001, 1995, 2003, 2009, 1993, 2010, 1994, 1998, 2002, 2006, 1997, 2005, 1990, 10, 2008, 2007, 2004, 1983, 1991...
}
\\
\midrule
\vspace{2pt}
\parbox[t]{2mm}{\rotatebox[origin=c]{90}{\textbf{d-RNN}}}
&
\pbox{7.7cm}{
\textbf{Examples}:
\textit{
``aortic root is dilated ( measured \textbf{37} x \textbf{37} mm'',
``ascending aorta is not dilated ( \textbf{32} x \textbf{31} mm''
} \\
\textbf{Numerals}: 42, 33, 31, 43, 44, 21, 38, 36, 46, 37, 32, 39, 26, 28, 23, 29, 45, 40, 49, 94...
}
&
\pbox{6cm}{
\textbf{Examples}:
\textit{
``ngc \textbf{6334} stars'',
``ngc \textbf{2366} shows a wealth of small structures''
}
\textbf{Numerals}: 294, 4000, 238, 6334, 2363, 1275, 2366, 602, 375, 1068, 211, 6.4, 8.7, 600, 96, 0.65, 700, 1.17, 4861, 270...
}
\\
\midrule
\parbox[t]{2mm}{\rotatebox[origin=c]{90}{\textbf{MoG}}}
&
\pbox{7.7cm}{
\textbf{Examples}:
\textit{
``stroke volume \textbf{46.1} ml'',
``stroke volume \textbf{65.6} ml'',
``stroke volume \textbf{74.5} ml'',
``end diastolic volume \textbf{82.6} ml'',
``end diastolic volume \textbf{99.09} ml'',
``end diastolic volume \textbf{138.47} ml''
}\\
\textbf{Numerals}: 74.5, 69.3, 95.9, 96.5, 72.5, 68.6, 82.1, 63.7, 78.6, 69.6, 69.5, 82.2, 68.3, 73.2, 63.2, 82.6, 77.7, 80.7, 70.7, 70.4...
}
&
\pbox{6cm}{
\textbf{Examples}:
\textit{
``hip \textbf{12961} and gl \textbf{676} a are orbited by giant planets,''
``velocities of gl \textbf{676}'',
``velocities of hip \textbf{12961}''
}\\
\textbf{Numerals}: 12961, 766, 7409, 4663, 44.3, 1819, 676, 1070, 5063, 323, 264, 163296, 2030, 77, 1.15, 196, 0.17, 148937, 0.43, 209458...
}
\\
\bottomrule
\end{tabular}
\end{center}
\caption{Examples of numerals with highest probability in each strategy of the combination model.}
\label{tab:most_discrete_nums}
\end{table*}

\paragraph{Selection of Strategy in Combination Model}

Table~\ref{tab:most_discrete_nums} shows development set examples with high selection probabilities for each strategy of the combination model, along with numerals with the highest average selection per mode. The h-softmax model is responsible for mostly integers with special functions, e.g. years, typical drug dosages, percentile points, etc.
In the clinical data, d-RNN picks up two-digit integers (mostly dimensions) and MoG is activated for continuous attributes, which are mostly out of vocabulary.
In the scientific data, d-RNN and MoG showed affinity to different indices from catalogues of astronomical objects: d-RNN mainly to NGC~\cite{dreyer1888new} and MoG to various other indices, such as GL~\cite{gliese1988third} and HIP~\cite{perryman1997hipparcos}.
In this case,
MoG was wrongly selected for numerals with a labelling function,
which also highlights a limitation of evaluating on the number line,
when a numeral is not used to represent its magnitude.

\begin{figure}[t]
\centering
\includegraphics[width=0.98\columnwidth]{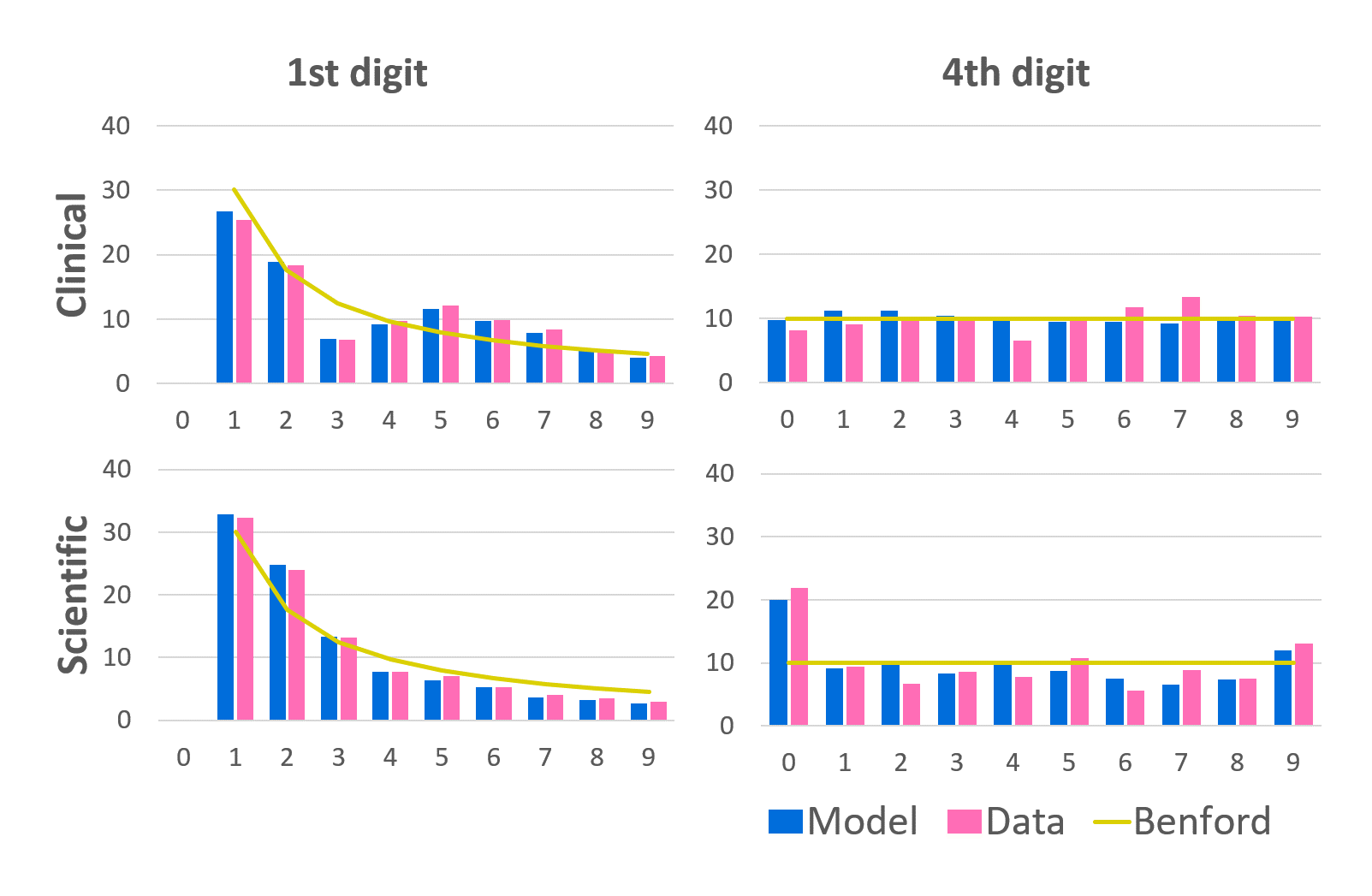}
\caption{Distributions of significant digits from d-RNN model, data, and theoretical expectation (Benford's law).}
\label{fig:benford}
\end{figure}
\paragraph{Significant Digits}
Figure~\ref{fig:pdf} shows the distributions of the most significant digits under the d-RNN model and from data counts. The theoretical estimate has been overlayed, according to Benford's law~\cite{benford1938law}, also called the first-digit law, which applies to many real-life numerals.
The law predicts that the first digit is 1 with higher probability (about 30\%) than 9 ($<5$\%) and weakens towards uniformity at higher digits.
Model probabilities closely follow estimates from the data.
Violations from Benford's law can be due to rounding~\cite{beer2009terminal} and can be used as evidence for fraud detection~\cite{DBLP:conf/ai/LuBC06}.

\section{Related Work}


Numerical quantities have been recognised as important for textual entailment~\cite{lev2004solving,dagan2013recognizing}.
\newcite{roy2015reasoning} proposed a quantity entailment sub-task that focused on whether a given quantity can be inferred from a given text and, if so, what its value should be.
A common framework for acquiring common sense about numerical attributes of objects has been to collect a corpus of numerical values in pre-specified templates and then model attributes as a normal distribution~\cite{aramaki2007uth,davidov2010extraction,ifteneuaic,narisawa2013204,de2010good}.
Our model embeds these approaches into a LM that has a sense for numbers.

Other tasks that deal with numerals are numerical information extraction and solving mathematical problems. Numerical relations have at least one argument that is a number and the aim of the task is to extract all such relations from a corpus, which can range from identifying a few numerical attributes~\cite{nguyen2011end,intxaurrondo2015diamonds} to generic numerical relation extraction~\cite{hoffmann2010learning,madaan2016numerical}.
Our model does not extract values, but rather produces an probabilistic estimate.

Much work has been done in solving arithmetic~\cite{mitra2016learning,hosseini2014learning,roy2016solving}, geometric~\cite{seo2015solving}, and algebraic problems~\cite{zhou2015learn,koncel2015parsing,upadhyay2016learning,upadhyay2016annotating,shi2015automatically,kushman2014learning} expressed in natural language.
Such models often use mathematical background knowledge, such as linear system solvers. The output of our model is not based on such algorithmic operations, but could be extended to do so in future work.

In language modelling, generating rare or unknown words has been a challenge, similar to our unknown numeral problem.
\newcite{gulcehre2016pointing} and \newcite{gu2016incorporating} adopted pointer networks~\cite{vinyals2015pointer} to copy unknown words from the source in translation and summarisation tasks. \newcite{merity2016pointer} and \newcite{lebret2016neural} have models that copy from context sentences and from Wikipedia's infoboxes, respectively.
\newcite{ahn2016neural} proposed a LM that retrieves unknown words from facts in a knowledge graph. They draw attention to the inappropriateness of perplexity when OOV-rates are high and instead propose an adjusted perplexity metric that is equivalent to APP.
Other methods aim at speeding up LMs to allow for larger vocabularies~\cite{chen2015strategies}, such as hierarchical softmax~\cite{morin2005hierarchical}, target sampling~\cite{jean2014using}, etc., but still suffer from the unknown word problem.
Finally, the problem is resolved when predicting one character at a time, as done by the character-level RNN~\cite{graves2013generating,sutskever2011generating} used in our d-RNN model.



\section{Conclusion}

In this paper, we investigated several strategies for LMs to model numerals and proposed a novel open-vocabulary generative model based on a continuous probability density function.
We provided the first thorough evaluation of LMs on numerals on two corpora, taking into account their high out-of-vocabulary rate and numerical value (magnitude).
We found that modelling numerals separately from other words through a hierarchical softmax can substantially improve the perplexity of LMs, that different strategies are suitable for different contexts, and that a combination of these strategies can help improve the perplexity further. Finally, we found that using a continuous probability density function can improve prediction accuracy of LMs for numbers by substantially reducing the mean absolute percentage metric.

Our approaches in modelling and evaluation can be used in future work in tasks such as approximate information extraction, knowledge base completion, numerical fact checking, numerical question answering, and fraud detection.
Our code and data are available at: \url{https://github.com/uclmr/numerate-language-models}.

\section*{Acknowledgments}
The authors would like to thank the anonymous reviewers for their insightful comments
and also Steffen Petersen for providing the clinical dataset and advising us on the clinical aspects of this work. This research was supported by the Farr Institute of Health Informatics Research and an Allen Distinguished Investigator award.

\bibliography{bibliography}
\bibliographystyle{acl_natbib}

\end{document}